\documentclass[11pt]{article}
\usepackage{coling2020}

\usepackage{url}
\usepackage{latexsym}
\usepackage{times}
\usepackage{mathtools}
\usepackage{array} 
\usepackage{booktabs}  
\usepackage{threeparttable}  
\usepackage{multicol}  
\usepackage{multirow}  
\usepackage{amsmath}
\usepackage{amsfonts}
\usepackage{epsfig}
\usepackage{algorithm}
\usepackage{algorithmic}
\usepackage{graphicx} 

\usepackage{microtype}
\usepackage{xspace}
\usepackage{xcolor}

\newcommand{\BERTBASE}{BERT$_{\small \textsc{BASE}}$\xspace}
\newcommand{\BERTLARGE}{BERT$_{\small \textsc{LARGE}}$\xspace}

\title{ MS-Ranker: Accumulating Evidence from Potentially Correct Candidates for Answer Selection
}

\author{Yingxue Zhang\textsuperscript{\rm 1}, Fandong Meng\textsuperscript{\rm 1}, Peng Li\textsuperscript{\rm 1}, Ping Jian\textsuperscript{\rm 2}, Jie Zhou\textsuperscript{\rm 1} \\ 
\textsuperscript{\rm 1}Pattern Recognition Center, WeChat AI, Tencent Inc, China\\ 
\textsuperscript{\rm 2}Beijing Institute of Technology, China\\

\{yxuezhang, fandongmeng, patrickpli, withtomzhou\}@tencent.com, 
pjian@bit.edu.cn\\
}

\date{}

\begin{document}
\maketitle
\begin{abstract}
As conventional answer selection (AS) methods generally match the question with each candidate answer independently, they suffer from the lack of matching information between the question and the candidate. To address this problem, we propose a novel reinforcement learning (RL) based multi-step ranking model, named MS-Ranker, which accumulates information from potentially correct candidate answers as extra evidence for matching the question with a candidate. In specific, we explicitly consider the potential correctness of candidates and update the evidence with a gating mechanism. Moreover, as we use a listwise ranking reward, our model learns to pay more attention to the overall performance. Experiments on two benchmarks, namely WikiQA and SemEval-2016 CQA, show that our model significantly outperforms existing methods that do not rely on external resources. 

\end{abstract}

\section{Introduction}
\noindent Answer Selection (AS) is the task of selecting correct answers to a given question from an answer candidate set, which is an active research field in recent years. Typically, AS is treated as a question-candidate matching problem, and various methods have been proposed to measure the similarity between a question and a candidate, including feature engineering based methods~\cite{wang2007jeopardy,yih2013question} and deep learning based methods~\cite{Parikh2016ADA,yin2016abcnn}.

Despite the fact that existing methods have achieved promising results, they still suffer from the lack of matching information between the question and the candidate. Taking Table~\ref{real_case} as an example, as both the candidates $C_2$ and $C_3$ share no information with the question, it would be difficult (if not impossible) to tell whether $C_2$ and $C_3$ are correct. Meantime, $C_1$ shares ``Life insurance'' with the question, which makes it a potentially correct answer. Suppose we utilize the information of $C_1$ as additional evidence, it would be easier to determine $C_2$ is also correct as there is a significant information overlap between $C_1$ and $C_2$. Therefore, we argue that explicitly considering the evidence carried by candidates (especially correct ones) in an AS model is necessary.

However, it is non-trivial to accumulate reliable cross-candidate evidence. A simple and intuitive way is to imitate the way how humans select answers: address one candidate a time, and determine its correctness by comparing it with both the question and the evidence accumulated from the already selected candidates. In this way, AS is reduced to a Markov decision process, and the model would be capable of only aggregating the information from potentially correct candidates, making the evidence more reliable and effective. Unfortunately, the above way introduces several discrete operations, which makes the model not end-to-end trainable.

\begin{table}[t]
\centering
\small
\begin{tabular}{p{0.02\columnwidth}p{0.72\columnwidth}l}
\toprule
\multicolumn{3}{l}{{\bf Question}: What does \underline{\textbf{life insurance}} cover?}\\
\midrule
  $C_1$ & \underline{\textbf{Life insurance}} is a contract between an insured (insurance \textit{\textbf{{policy}}} holder) and an insurer, where the insurer promises to \textit{\textbf{{pay}}} a designated \textit{\textbf{{beneficiary}}} a sum of \textit{\textbf{{money (the ``benefits'')}}} upon the \textit{\textbf{{death}}} of the insured person. & True\\  
  $C_2$ & Depending on the contract, other events such as \textit{\textbf{{terminal illness}}} or \textit{\textbf{{critical illness}}} may also trigger \textit{\textbf{{payment}}}.  & True\\
  $C_3$ & Life-based contracts tend to fall into two major categories & False \\
  $C_4$ & Protection \textit{\textbf{{policies}}} - designed to provide a \textit{\textbf{{benefit}}} in the event of specified event, typically a lump sum \textit{\textbf{{payment}}}. & True\\
\bottomrule
\end{tabular} 
\caption{A real case from WikiQA. To save space, we have omitted a few incorrect candidates. The bold italic text is the information carried by correct candidates which can be evidence along with the question.
} 
\label{real_case}
\end{table}

To alleviate the above problems, we propose a reinforcement learning (RL)~\cite{williams1992simple} based multi-step ranking model named as MS-Ranker for AS, which is capable of accumulating reliable evidence from candidates and end-to-end trainable. To be specific, the method consists of a Pre-Ranker and a RL agent. The Pre-Ranker is used to pre-rank all candidates independently to determine the order they will be examined by the RL agent, aiming to make the potentially correct candidates (such as $C_1$ in Table~\ref{real_case}) be examined earlier and provide evidence for later decisions. The RL agent re-ranks all candidates one by one and accumulates evidence from potentially correct candidates step by step. At each step, cross-candidate evidence is incorporated into the model in two ways: (1) When ranking one candidate, the RL agent compares the candidate with both the question and the evidence to assign a more accurate correctness score; (2) After ranking the current candidate, the RL agent integrates its information into the evidence according to the assigned score with a gating mechanism. The new evidence will be used in the next step. In addition, a listwise ranking reward is leveraged in RL training, which makes the agent pay more attention to the whole picture. As a by-product of RL, our model is end-to-end trainable although there are discrete operations. Our major contributions \footnote{We will release the code at GitHub upon publication.} are as follows:
\vspace{-7pt}
\begin{itemize}
    \item To the best of our knowledge, we are the first to  explicitly accumulate evidence from candidates based on their potential correctness on the answer selection task. 
    \vspace{-7pt}
    \item We propose a novel RL based MS-Ranker for answer selection, which leverages a listwise ranking reward to make the model pay more attention to the overall ranking performance.
    \vspace{-7pt}
    \item Experiments on two AS datasets WikiQA and SemEval-2016 CQA show that our model gains considerable improvements over existing systems that do not rely on external resources.
\end{itemize}

\section{Related Work}
\subsection{Answer Selection}
Answer selection (AS) has attracted intensive attention in recent years. Earlier methods are based on syntactic or semantic features~\cite{wang2007jeopardy,heilman2010tree,yih2013question} or structural kernels~\cite{severyn2012structural,Severyn2013BuildingSF,tymoshenko2014encoding}, which require lots of sophisticated feature engineering work and suffer from data sparsity problem. Most recent AS models are based on deep neural networks, including attentive networks~\cite{Parikh2016ADA,yin2016abcnn,wang2016inner,sha2018multi} and compare-aggregate networks~\cite{Wang2017ACM,Bian2017ACM}. These approaches typically model each pair of question-candidate independently and suffer from the information gap between the question and the candidate. 

Some researchers exploit utilizing additional information to improve the performance of answer selection. Yoon et al.~\shortcite{Yoon2019ACM} aggregate all the samples of the whole corpus into several clusters and utilize cluster information to help make decisions. Recently, many studies focus on transferring knowledge from external resources to improve ranking, such as external text corpora~\cite{savenkov2017evinets}, constituency or dependency tree~\cite{tymoshenko2018cross}. Li et al.~\shortcite{Li2018AUM} utilize large-scale English Wikipedia corpus to pre-train the LDA and LSA to improve answer selection. Recently, researches~\cite{lai2019gated,garg2019tanda} adopt transfer learning techniques on the pre-trained transformer models like BERT. They firstly fine-tune the pre-trained model with a large-scale external dataset, then perform a second fine-tuning step to adapt the target dataset. Different from these work, we utilize the natural correlation between candidates, referring other candidates which are more evident and easily detectable to help make a correct decision.

\subsection{Cross-Candidate Evidence and RL}
Recently, some work in open-domain QA~\cite{Wang2018R3RR,Lin2018DenoisingDS,Das2019MultistepRI} and machine reading comprehension (MRC)~\cite{Min2019MultihopRC,yu2019inferential} also exploit cross-candidate evidence and the application of RL. Our work is different from them since there are some key differences between Answer Selection (AS) and Open-domain QA or MRC. Specifically, the candidates in Open-domain QA or MRC are always selected from large corpora of text and are lack of annotations, while candidates in AS have been annotated according to their correctness. Our MS-Ranker is designed for AS, which effectively utilizes the unique annotation information of candidates to guide the accumulation of cross-candidate evidence. This annotation plays important roles in two aspects: (1) When gathering the evidence, we utilize the predicted correctness score to control how much information of the candidate should be added into the evidence, and the predicted score can be supervised by the annotations. (2) We design listwise ranking rewards that are computed based on the annotations of all candidates, which can guide the accumulation of evidence through the overall ranking performance. 

\section{Methodology}
In this section, we will introduce the detail of our proposed novel AS model MS-Ranker.

\begin{figure*}[t!]
\centering
\includegraphics[width=0.95\textwidth]{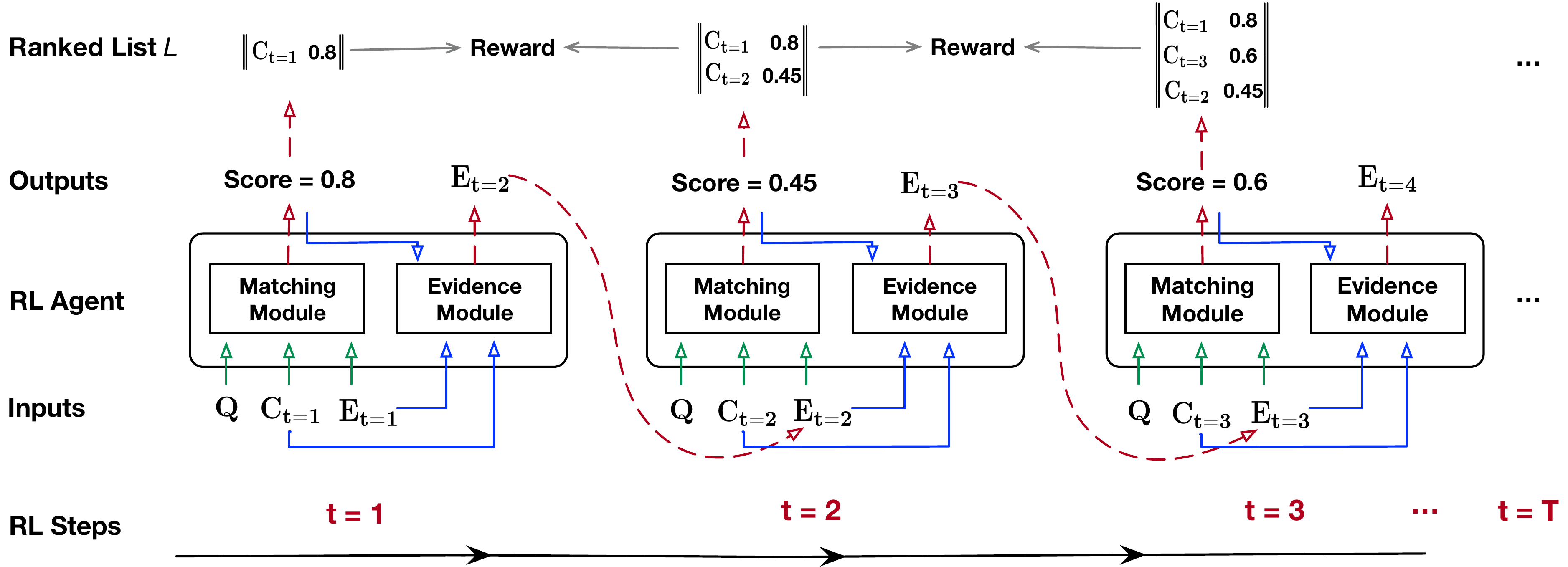}
\caption{The way RL agent works. The agent consists of the Matching Module and the Evidence Module, and it ranks one candidate a time. At step $t$ ($1\leq t\leq T$), the inputs of the Matching Module (green lines) are question $Q$, the current candidate $C_t$ and the current evidence $E_t$, and it assigns a correctness score for $C_t$. The inputs of the Evidence Module (blue lines) are $C_t$, $E_t$ and the correctness score, and it generates $E_{t+1}$ for the next step. We use the representation of the question ($Q$) to initialize the first evidence $E_{t=1}$.
}
\centering
\label{model_architechture}
\end{figure*}

\subsection{Task Definition} 
Before diving into the details of our model, we give the definition of the AS task first. 
Given a question $Q$ and its candidate answer set $C = \{C_1, C_2,\cdots,C_T\}$, the answer selector (or the ranker) aims to pick out all correct answers from the set $C$, i.e., rank correct answers before incorrect ones. For each candidate, the ranker assigns a correctness score for it, which is used for ranking. 

\subsection{Overview}
Our MS-Ranker model consists of a \emph{Pre-Ranker} and a \emph{RL agent}. The Pre-Ranker is used to pre-rank all candidates to determine the orders they are examined by the RL agent. And the RL agent is used to accumulate evidence and re-rank candidates.

\subsubsection{Pre-Ranker} The RL agent examines the candidates one by one and accumulates evidence step by step accordingly. Intuitively, the order these candidates enter the agent will affect the aggregation of the evidence. Therefore, we design a Pre-Ranker, which assigns a preliminary correctness score for each candidate through a pretrained question-candidate matching network. The candidate with a higher score enters the RL agent earlier. In this way, the candidates sharing information with the question can be examined first and provide more reliable evidence for later steps. Specifically, the pretrained network is implemented as an encoding layer (Section~\ref{sec:encoding-layer}) and a question-candidate attention layer (Section~\ref{sec:attention-layer}) followed by an MLP layer.  
\subsubsection{RL Agent} As shown in Figure~\ref{model_architechture}, the agent maintains a ranked candidate list $L$. It selects one candidate at each step. Therefore, the number of candidates for a question determines the steps of the RL process. The agent consists of two modules: the Matching Module and the Evidence Module. At step $t$ ($1\leq t\leq T$, where $T$ is the number of candidates for the current question, which is also the number of RL steps), the agent firstly assigns a correctness score for candidate $C_t$ based on the question $Q$ and the evidence $E_t$ through the Matching Module (Section~\ref{base}). $C_t$ is then inserted into $L$ according to the score. After that, the agent updates the evidence to obtain $E_{t+1}$ for the next step through the Evidence Module (Section~\ref{evidence}). 

\subsection{Matching Module}
\label{base}
\begin{figure}[t!]
    \centering
    \includegraphics[width=7.8cm]{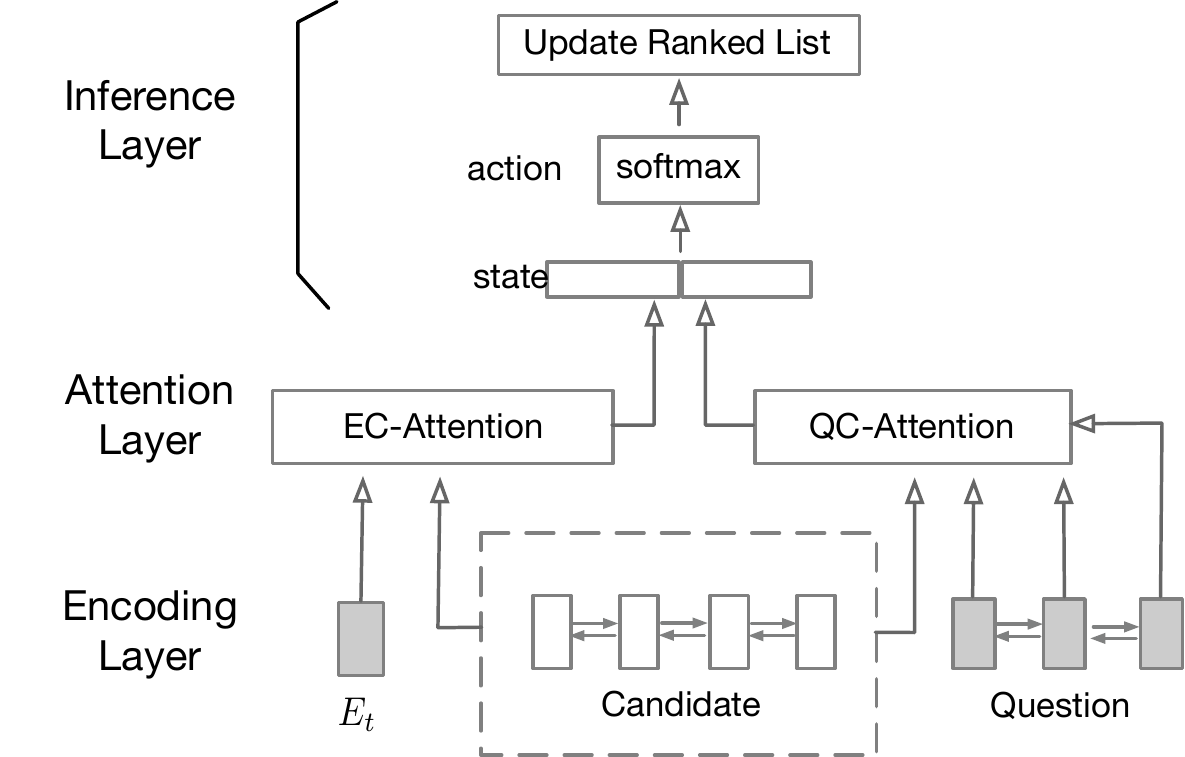}
    \caption{Matching Module. It computes two attentions. The QC-Attention is for matching the question and the candidate, while the EC-Attention is for matching the evidence and the candidate.
    } 
    \vspace{-6pt}
    \label{matching}
\end{figure}

The Matching Module assigns a correctness score for each candidate via comparing it with both the question and the current evidence. As shown in Figure~\ref{matching}, the Matching Module consists of three layers, the encoding layer, the attention layer, and the inference layer.

\subsubsection{Encoding Layer}
\label{sec:encoding-layer}
At each step, the question $Q$ and the current candidate $C$ is encoded into vectors by a one-layer Bidirectional Gated Recurrent Unit network (BiGRU)~\cite{Cho2014LearningPR}. 
Firstly, $Q$ and $C$ are represented as $X_Q=(x_{q_1},x_{q_2},...,x_{q_{|Q|}}) \in \mathbb{R}^{|Q|\times d_e}$ and $X_C=(x_{c_1},x_{c_2},...,x_{c_{|C|}})\in \mathbb{R}^{|C|\times d_e}$, respectively, where $x_{q_i}$ and $x_{c_j}$ are words whose embeddings are initialized with pre-trained $d_e$-dimension word embeddings. $|Q|$ and $|C|$ represent the length of the question and the candidate answer, respectively. Then, we feed $Q$ and $C$ into the BiGRU:
\begin{eqnarray} 
    h_{q_i} &=& \texttt{BiGRU}(X_Q,i), \\
    h_{c_j} &=& \texttt{BiGRU}(X_C,j) ,
\end{eqnarray}
where $h_{q_i}$ and $h_{c_j}$ indicate the hidden states of the \textit{i}-th word in $Q$ and the \textit{j}-th word in $C$, respectively. Then we get $H_Q=(h_{q_1}, h_{q_2},...,h_{q_{|Q|}})\in \mathbb{R}^{|Q|\times2d_g}$ and $H_C={(h_{c_1}, h_{c_2},...,h_{c_{|C|}})}\in  \mathbb{R}^{|C|\times2d_g}$ as the contextual representations of $Q$ and $C$, where $d_g$ is the hidden size of the GRU.

\subsubsection{Attention Layer}
\label{sec:attention-layer}
In this layer, in order to cover important information of both the question and the evidence, we compute two attentions. The one is the question-candidate attention (QC-attention), which is responsible for extract question-candidate matching information.
The other is the evidence-candidate attention (EC-attention), which can utilize the accumulated evidence to further match the candidate.
We adopt the attention mechanism from the Bi-Directional Attention Flow (BiDAF) model~\cite{Seo2017BidirectionalAF} 

\paragraph{QC-Attention.} We firstly compute question-aware vector $U_{c_j}$ for words of the candidate as follows:
\begin{eqnarray}
    \alpha_{ij}&=&  v_1 h_{q_i} + v_2 h_{c_j} + v_3(h_{q_i} \odot h_{c_j}),\\
    p_{ij}& = & e^{\alpha_{ij}} \bigg/ \sum_{i=1}^{|Q|} e^{\alpha_{ij}},\\
    U_{c_j} & = & \sum_{i=1}^{|Q|}p_{ij}h_{q_i},
\end{eqnarray}
where $v_1\in \mathbb{R}^{2d_g}$, $v_2\in \mathbb{R}^{2d_g}$ and $v_3\in \mathbb{R}^{2d_g}$ are parameters and $\odot$ stands for the element-wise multiplication. The attention representation of the candidate is $U_C = \{U_{c_1},U_{c_2},\cdots,U_{c_{|C|}}\} \in \mathbb{R}^{|C|\times2 d_g}$. Then we encode the question into a candidate-aware vector $U_Q$ as follows:
\begin{eqnarray}
    \beta_j &=& \max \limits_{1\leq i\leq |Q|}\{\alpha_{ij}\},\\
    p_{j} &=& e^{\beta_j} \bigg/ \sum_{j=1}^{|C|} e^{\beta_j},\\
    U_Q &=& \sum_{j=1}^{|C|} p_{j} h_{c_j}.
\end{eqnarray}
The attention representation of the question is $U_Q \in \mathbb{R}^{2d_g}$. We then compute $m_j$ as follows:
\begin{equation}
    m_j=[h_{c_j};U_{c_j};h_{c_j} \odot U_{c_j}; U_Q\odot U_{c_j}],
\end{equation}
where $m_j \in \mathbb{R}^{8d_g}$ is the matching representation between the question and the \textit{j}-th token of the candidate. We then feed the QC-matching matrix 
    $M_{qc}=\{m_1,m_2,\cdots,m_{|C|}\} \in \mathbb{R}^{|C|\times8d_g}$
into a one-layer MLP followed by a max-pooling layer to obtain the final QC-matching representation $V_{qc}\in \mathbb{R}^{2d_g}$:
\begin{eqnarray}
    \tilde M_{qc} &=& \tanh(W_{qc}M_{qc}+b_{qc}),\\
    V_{qc} &=& \texttt{MaxPool}(\tilde M_{qc}),
\end{eqnarray}
where $W_{qc} \in \mathbb{R}^{2d_g\times8d_g}$ and $b_{qc} \in \mathbb{R}^{2d_g}$ are parameters.

\paragraph{\bf EC-Attention.}
The evidence is always a vector (Section~\ref{evidence}). At each step, given the evidence vector $E\in \mathbb{R}^{2d_g}$ and the contextual representation of the current candidate $H_{C}={(h_{c_1}, h_{c_2},...,h_{c_{|C|}})}\in \mathbb{R}^{|C|\times2d_g}$, we obtain their matching representation $U_E \in \mathbb{R}^{2d_g}$:
\begin{eqnarray}
    \alpha_{j}&=&v_4 E + v_5 h_{c_j} + v_6(E \odot h_{c_j}), \\
    p_{j} &=& e^{\alpha_{j}} \bigg/ \sum_{j=1}^{|C|} e^{\alpha_{j}},\\
    U_E &=& \sum_{j=1}^{|C|} p_{j}h_{c_j},
\end{eqnarray}
where $v_4\in \mathbb{R}^{2d_g}$, $v_5\in \mathbb{R}^{2d_g}$, and $v_6\in \mathbb{R}^{2d_g}$ are parameters. Then the matching vector $u_j$ between tokens of candidate and evidence vector is compute as:
\begin{equation}
    u_j= [h_{c_j}; h_{c_j} \odot U_E],
\end{equation}
where $u_j \in \mathbb{R}^{4d_g}$ represents the matching vector between the \textit{j}-th token of the candidate and the question. We then feed the E-to-C matching matrix
    $M_{ec}=\{u_1,u_2,...,u_{|C|}\} \in \mathbb{R}^{|C|\times 4d_g}$
into a one-layer MLP followed by a max-pooling layer to obtain the final EC-matching representation $V_{ec}\in \mathbb{R}^{2d_g}$:
\begin{eqnarray}
    \tilde M_{ec} &=& \tanh(W_{ec}M_{ec}+b_{ec}),\\
    V_{ec} &=& \texttt{MaxPool}(\tilde M_{ec}),
\end{eqnarray}
where $W_{ec} \in \mathbb{R}^{2d_g\times4d_g}$ and $b_{ec} \in \mathbb{R}^{2d_g}$ are parameters.

\subsubsection{Inference Layer}
\label{infer}
Suppose we are at step $t$, then we obtain the RL state as $ s_t = [V_{qc}; V_{ec}]$, which is the concatenation of  $V_{qc}$ and $V_{ec}$. The agent samples an action $a_t$ based on the state $s_t$. In our method, the action space is defined as \{0, 1\}, where 1 indicates the current candidate can answer the question while 0 indicates it cannot. Then the correctness score $P_{pos_t}$ used for ranking is the value of $p(a_t=1|s_t)$. Specifically, the agent maps the state to a probability distribution over all possible actions through an MLP as follows:
\begin{eqnarray}
    f(s_t) &=& \tanh(W_1s_t+b_1),\\
    p(a_t|s_t) &=& \mathrm{softmax}(W_2f(s_t)+b_2),\\
    P_{pos_t}&=& p(a_t=1|s_t).
\end{eqnarray}
The agent inserts the current candidate into the ranking list $L$ based on the score $P_{pos_t}$.

\subsection{Evidence Module}
\label{evidence}
The Evidence Module aims to accumulate evidence from historical reliable candidates. For the first step, since there is no historical candidates before, the evidence $E_1$ is initialized as the representation of the question. Specifically, we feed the question into a one-layer BiGRU and concatenate the last hidden states of the BiGRU in two directions as the first evidence $E_1 \in \mathbb{R}^{2{d_g}}$.

Then at following steps, the agent integrates the information of reliable candidates into the evidence step by step. Suppose we are at step $t$, the agent has assigned a score $P_{pos_t}$ for the candidate $C_t$. If $P_{pos_t}$ is smaller than the threshold (We set the threshold as 0.5, which is determined over the development set), the information of $C_t$ will not be integrated into the evidence. We copy $E_t$ as the evidence $E_{t+1}$ for step $t+1$: $E_{t+1} = E_t$. Otherwise, if $P_{pos_t}$ is larger than the threshold, we view $C_t$ as a potentially correct answer, then the agent integrates its information into the current evidence $E_t$ to obtain $E_{t+1}$: $ E_{t+1} = \mathbb{F}(E_t, C_t)$.

Specifically, we design two ways to control the accumulation of the evidence. 

(1) {\bf RL Action}: To make sure that the noise from incorrect candidates will not affect the evidence a lot, we utilize $P_{{pos}_t} = P(a_t=1|s_t)$ (RL action) to decide how much information of $C_t$ should be added into the evidence. Firstly, $C_t$ is encoded into vector by BiGRU to get $O_t$, where $O_t$ is the concatenation of the last hidden states of the BiGRU in two directions. Then, we get $\tilde{O}_{t}$ as follows:
\begin{eqnarray} 
    \tilde{O}_t &=& P_{{pos}_t} O_t,
\end{eqnarray} 
$\tilde{O}_{t}$ carries the information of $C_t$ which will be integrated into the new evidence. 

(2) {\bf Gating Mechanism}: To ensure that relevant information from $E_t$ is preserved and new information from $\tilde O_t$ is added to $E_{t+1}$, we utilize the gating mechanism to control the information flow:
\begin{eqnarray}
    g &=& \sigma\left(W_e E_t + W_o \tilde O_t\right),\\
    E_{t+1} &=& (1-g) \odot E_t + g \odot \tilde O_t,
\end{eqnarray}
where $\sigma(\cdot)$ is a sigmoid function. $W_e$, $W_o$ are parameters, $g$ is the gate to control information flow. The evidence $E_{t+1}$ will be used for predicting the next candidate $C_{t+1}$.

\subsection{Training}
\paragraph{Listwise Ranking Rewards.} The model is optimized with obtained rewards at each time step. Since the task is modeled as a sequential ranking problem, we design the reward based on a commonly used metric for ranking, Average Precision (AveP), which is a listwise metric. Specifically, after the agent predicting $P_{pos}$, the current candidate is added into the ranking list $L$, which determines the value of AveP. Intuitively, if the value of AveP remains the same, which means the model predicts a correct action for a wrong candidate and adds it to the lower positions in the rank, we assign a small positive rewards 0.1 for it. Otherwise, we calculate the reward as the difference of the AveP before and after the change. If the value of AveP gains improvement, which means the current decision improves overall ranking performance, then the agent will obtain a positive reward. Instead, it will receive a negative reward. We use $AP_t$ to represent the AveP of $L$ at step t,  $AP_{t-1}$ to represent the AveP of $L$ at step $t-1$. The function of the reward $R(a_{1:T})$ is as follows:
\vspace{-8pt}
\begin{equation}
 R(a_t)=
    \begin{cases}
        0.1 & AP_t = AP_{t-1}\\
        AP_t - AP_{t-1} & AP_t \neq AP_{t-1}
    \end{cases},
\end{equation}
\begin{equation}
    AP_t = \frac{1}{N}\sum_{n=1}^N \frac{n}{position(n)} ,
\end{equation}
where $N$ represents the number of correct candidates (in ground-truth) in the first $t$ candidates. ``$position(n)$'' stands for the ranked position of the $n$-th correct candidate answer in $L$ at step $t$. 

\paragraph{Objective Function} In this work, we optimize the parameters of the policy network using REINFORCE algorithm~\cite{williams1992simple}, which aims to maximize the expected reward:
\begin{equation}
J(\theta)= \mathbb{E}_{a_{1:T}\sim p_\theta(a_t|s_t)}R(a_{1:T}),
\end{equation}
and approximate the gradient via sampling as
\begin{equation}
\nabla_\theta J(\theta)\simeq \sum_{t=1}^TR(a_t)\nabla_\theta\log p_\theta(a_t|s_t).
\end{equation}
\vspace{-20pt}
\section{Experiments}

\subsection{Datasets}
We conduct experiments on two datasets:

(1)  \textbf{WikiQA}~\cite{yang2015wikiqa} is an answer selection dataset constructed from real questions of Bing and Wikipedia. We remove all questions with no correct candidate answers as predecessors~\cite{yang2015wikiqa,Wang2017ACM} do, then the train/dev/test set contains 873/126/243 questions and 8627/1130/2351 question-candidate pairs, respectively.


(2) \textbf{SemEval-2016 CQA}~\cite{Nakov2016SemEval2016T3} is the dataset of SemEval-2016 Task 3: Community Question Answering. We focus on Subtask A, and the train/dev/test set contains 4873/244/327 questions and 36191/2440/3270 question-candidate pairs, respectively.



\subsection{Baselines}
We compare MS-Ranker with various baselines, and all the models are evaluated with the commonly used metrics Mean Average Precision (MAP) and Mean Reciprocal Rank (MRR).
   
(1) \textbf{ABCNN}~\cite{yin2016abcnn}: An attention-based CNN.
    
(2) \textbf{INRNN}~\cite{wang2016inner}: A RNN model that adds attention information.
    
(3) \textbf{IWAN-skip}~\cite{shen2017inter}: A method that discover fine-grained alignment of two sentences.
   
(4) \textbf{CA-network}~\cite{Wang2017ACM}: A general ``compare-aggregate'' framework.
    
(5) \textbf{MVFNN}~\cite{sha2018multi}: A Multi-View Fusion Neural Network with 4 attention modules.
    
(6) \textbf{Kernel}~\cite{tymoshenko2018cross}: A kernel based method that designs lots of linguistic features, which requires intricate feature engineering work and uses many external resources. 
    
(7) \textbf{HyperQA}~\cite{tay2018hyperbolic}: A ranking method based on embeddings in Hyperbolic space.
   
(8) \textbf{LC+ELMo}~\cite{Yoon2019ACM} A model with latent clustering and transfer learning. 

(9) \textbf{\BERTBASE+GSAMN}~\cite{lai2019gated}: A new gated self-attention memory network based on \BERTBASE. We list their results obtained without using a large-scale external dataset for a fair comparison.

\subsection{Experimental Settings}
We initialize word embeddings with $300$-dimensional GloVe vectors~\cite{pennington2014glove}. A mini-batch contains $10$ questions and the corresponding candidates. We set the hidden size of GRUs to $128$, the dropout rate to $0.5$, the learning rate to $10^{-3}$ which is decayed after every epoch by a factor of $0.99$. The number of training epochs for the Pre-Ranker is $5$, and the encoding layer and attention layer of the RL agent are initialized with those of the Pre-Ranker and fine-tuned.

\begin{table}[t]
\begin{minipage}{0.5\linewidth}
\centering
\scalebox{0.75}
{
\begin{tabular}{lcccc} 
\toprule
 \multicolumn{1}{c}{\multirow{2}*{\textbf{Model}}} & \multicolumn{2}{c}{\textbf{WikiQA}} & \multicolumn{2}{c}{\textbf{SemEval-2016}}\\
\cmidrule(r){2-3}\cmidrule(r){4-5}
& \textbf{MAP} & \textbf{MRR}& \textbf{MAP} & \textbf{MRR}\\ 
\midrule
ABCNN&69.21&71.28&~~75.79$^*$&~~81.43$^*$\\
INRNN&73.41&74.18&-&-\\
IWAN-skip&73.30&75.00&-&-\\
CA-network&74.33&75.45&~79.05$^*$&~~86.22$^*$\\
MVFNN&74.62&75.76&80.05&87.18\\
Kernel&\textbf{75.29}&76.21&79.79&86.52\\
HyperQA&71.20&72.70&79.50&-\\
\midrule
MS-Ranker (Ours) &75.04&\textbf{76.56}&\textbf{80.71}&\textbf{88.04}\\
\bottomrule
\end{tabular}}
\caption{Results on WikiQA and SemEval-2016. 
}
\vspace{-15pt}
\label{Wiki_results}
\end{minipage}\begin{minipage}{0.5\linewidth}  
\centering
\scalebox{0.8}
{\begin{tabular}{lcc} 
\toprule
 \multicolumn{1}{l}{\multirow{1}*{\textbf{Model}}} 
& \multicolumn{1}{c}{\textbf{MAP}} & \multicolumn{1}{c}{\textbf{MRR}}\\ 
\midrule
MS-Ranker (ours)&75.04&76.56\\
LC + ELMo \cite{Yoon2019ACM}&76.40&\textbf{78.40}\\
MS-Ranker + ELMo&\textbf{76.51}&77.23\\
\midrule
\BERTBASE+GSAMN~\cite{lai2019gated} &82.10&83.20\\
\BERTBASE &81.29&82.46\\
\BERTBASE + MS-Ranker&\textbf{82.25}&\textbf{83.71}\\
\midrule
\BERTLARGE &83.31&84.27\\
\BERTLARGE + MS-Ranker&\textbf{84.14}&\textbf{84.98}\\
\bottomrule
\end{tabular} }
\caption{Results on language models of WikiQA. 
}
\vspace{-15pt}
\label{LM_results}
\end{minipage}
\end{table}

\subsection{Main Results}
\subsubsection{WikiQA and SemEval-2016 CQA} Table~\ref{Wiki_results} shows the results on the benchmarks of WikiQA and SemEval-2016 CQA. On the non-factoid dataset SemEval-2016 CQA, our method gains improvements of $0.86\%$ on MRR and $0.66\%$ on MAP. On WikiQA, compared with Kernel~\cite{tymoshenko2018cross}, our approach obtains almost the same MRR while obtains slightly lower MAP, which mainly because of the following two reasons: (1) Kernel uses lots of external resources such as the constituency or dependency tree and WordNet~\cite{miller1995wordnet}, while we do not use any external resources; (2) Kernel carefully designs lots of   sophisticated features such that it is more suitable for the small, clean dataset WikiQA. Therefore, on the large, noisy dataset SemEval-2016 CQA, our approach performs better than Kernel ($+1.52\%$ on MRR, $+0.92\%$ on MAP).

\subsection{Experiments on Pre-trained Language Models}
We conduct experiments based on pre-trained language models to verify the effectiveness of our approach:

(1) \textbf{ELMo} ~\cite{Peters:2018}: We replace the GloVe embeddings with the ELMo embeddings. 

(2) \textbf{BERT}~\cite{Devlin2018BERTPO}: 
We replace the Pre-Ranker and the Encoding Layer with BERT (including \BERTBASE and \BERTLARGE). The QC-Attention Module is removed since BERT can support enough interaction between the question and the candidate. At the pre-rank stage, for \BERTBASE, we set the fine-tuning epochs as $5$, the learning rate as $2\times 10^{-5}$. For \BERTLARGE, we set the fine-tuning epochs as $4$, the learning rate as $1\times 10^{-5}$. Then, at the RL-rank stage, we set the learning rate as $1\times 10^{-5}$ for both the \BERTBASE and \BERTLARGE. The results on the development set are reported in Table~\ref{LM_results}. We can conclude that: (1) Utilizing pre-trained language models further improves the performance of our model. (2) MS-Ranker gains consistent improvements on strong BERT baselines, which reflects its generalization ability.


\subsection{Ablation Study}

In this section, we evaluate the impact of the main components of MS-Ranker on the validation set of WikiQA, including (1) Listwise Ranking Rewards, (2) Pre-Ranker, (3) Evidence, (4) RL Action, and (5) Gating Mechanism. We conduct comparative experiments to see how each technique works. Table~\ref{ablation} shows the ablation study results.

\subsubsection{Listwise Ranking Rewards (Row 2)}
We build a model without using the Listwise Ranking Rewards and compare it with the MS-Ranker. The model is trained with the loss function of cross-entropy. Row 1 v.s. Row 2 in Table~\ref{ablation} shows that the Listwise Ranking Rewards contributes significantly to the performance of MS-Ranker, resulting in a considerable increase.

\subsubsection{Pre-Ranker (Row 3)} 
We remove the Pre-Ranker from MS-Ranker to show its impact. Row 1 v.s. Row 3 in Table~\ref{ablation} shows that removing the Pre-Ranker results in a sharp decline in the performance. Without the Pre-Ranker, the agent will examine candidates in arbitrary order, making it more likely to accumulate information from incorrect candidates in the first few steps for further decisions, which makes the evidence unstable and brings the model large bias.

\subsubsection{Evidence (Row 4)} 
We build a model without using the evidence from candidates to prove the effectiveness of evidence accumulated in MS-Ranker. Specifically, we remove the Evidence Module and the EC-Attention Layer~(Section \ref{sec:attention-layer}) of the MS-Ranker to obtain the evidence-unaware model. Correspondingly, the correctness score is assigned only based on the question-candidate matching. Row 1 v.s. Row 4 in Table~\ref{ablation} shows that the MS-Ranker significantly outperforms the evidence-unaware model, which demonstrates that cooperating reliable evidence into AS models is necessary.

\begin{table} [t]
	\small
	\centering
	\begin{tabular}{llll}
		\toprule
		\multicolumn{1}{c}{\multirow{1}*{\textbf{\#}}}&
		\multicolumn{1}{c}{\multirow{1}*{\textbf{Model}}} 
	    & \multicolumn{1}{c}{\textbf{MAP}}& \multicolumn{1}{c}{\textbf{MRR}}\\ 
		\midrule
		1&MS-Ranker	&\textbf{75.17} &\textbf{76.36}\\
		2&~~~~w/o Rewards &74.31 (-0.86) &75.28 (-1.08) \\
		3&~~~~w/o Pre-Ranker&73.88 (-1.29)&74.79 (-1.57) \\
		4&~~~~w/o Evidence &73.43 (-1.74)&74.10 (-2.26)\\
		5&~~~~w/o RL action &74.20 (-0.97) & 75.21 (-1.15)\\
		6&~~~~w/o the Gate &74.51 (-0.65) & 75.45 (-0.91)\\
		\bottomrule
	\end{tabular}
	\caption{Results of ablation study on the validation set of WikiQA. 
	}
	
	\label{ablation}
\end{table}

We also conduct experiments to prove that the accumulation of evidence requires careful guidance and deliberation. To be specific, we utilize RL actions and the gating mechanism to guide the accumulation in the MS-Ranker and we verify their effectiveness in the following two experiments. 

\subsubsection{RL action  (Row 5)}
We remove the design of RL actions in the Evidence Module, i.e., evidence is accumulated without considering the potential correctness of each candidate. Row 1 v.s. Row 5 shows that removing the guidance of actions results in a significant quality drop, which proves that considering the potential correctness of each candidate is important when cooperating cross-candidate evidence into AS models. 

\subsubsection{Gating Mechanism  (Row 6)}To prove the effectiveness of the Gating Mechanism in Section~\ref{evidence}, we remove it and obtain $E_{t+1}$ as $E_{t+1} = \alpha E_t + (1-\alpha)  \tilde{O}_t$, where $\alpha = 0.5$. Row 1 v.s. Row 6 shows that utilizing the gating mechanism to control the evidence accumulation significantly boosts the performance.

\section{Conclusion}
In this work, we argue that the problem of scarce matching information between the question and the candidate is a common problem faced by most existing answer selection (AS) methods. To alleviate the problem, we propose a novel reinforcement learning (RL) based answer selection model named as MS-Ranker, which leverages an RL agent to collect reliable information from potentially correct candidates as extra evidence for question-candidate matching. A listwise ranking reward is also used to improve the overall ranking performance. Experiments on WikiQA and SemEval-2016 CQA show that our approach consistently outperforms the models which only consider question-candidate matching. 

\bibliographystyle{acl}
\bibliography{coling2020}

\end{document}